\documentclass[sigconf]{acmart}

\usepackage{booktabs} 
\usepackage{multirow}
\usepackage{array}
\usepackage{bm}

\setcopyright{rightsretained}

\begin{document}

\copyrightyear{2017}
\acmYear{2017}
\setcopyright{acmlicensed}
\acmConference{MM '17} {} {October 23--27, 2017, Mountain View, CA, USA}
\acmPrice{15.00}
\acmDOI{https://doi.org/10.1145/3123266.3123355}
\acmISBN{978-1-4503-4906-2/17/10}

\fancyhead{}
\settopmatter{printacmref=false, printfolios=false}

\title{Deep Binary Reconstruction for Cross-modal Hashing}
\author{Xuelong Li}
\affiliation{%
  \institution{OPTIMAL in Northwestern Polytechnical University}
  \streetaddress{127 West Youyi Road}
  \city{Xi'an}
  \state{Shaanxi, China}
}
\email{xuelong\_li@opt.ac.cn}

\author{Di Hu}
\affiliation{%
  \institution{OPTIMAL in Northwestern Polytechnical University}
  \streetaddress{127 West Youyi Road}
  \city{Xi'an}
  \state{Shaanxi, China}
}
\email{hdui831@mail.nwpu.edu.cn}

\author{Feiping Nie}
\authornote{Corresponding Author}
\affiliation{%
  \institution{OPTIMAL in Northwestern Polytechnical University}
  \streetaddress{127 West Youyi Road}
  \city{Xi'an}
  \state{Shaanxi, China}
}
\email{feipingnie@gmail.com}

\begin{abstract}
With the increasing demand of massive multimodal data storage and organization, cross-modal retrieval based on hashing technique has drawn much attention nowadays.
It takes the binary codes of one modality as the query to retrieve the relevant hashing codes of another modality.
However, the existing binary constraint makes it difficult to find the optimal cross-modal hashing function.
Most approaches choose to relax the constraint and perform thresholding strategy on the real-value representation instead of directly solving the original objective.
In this paper, we first provide a concrete analysis about the effectiveness of multimodal networks in preserving the inter- and intra-modal consistency.
Based on the analysis, we provide a so-called \emph{Deep Binary Reconstruction} (DBRC) network that can directly learn the binary hashing codes in an unsupervised fashion.
The superiority comes from a proposed simple but efficient activation function, named as \emph{Adaptive Tanh} (ATanh).
The ATanh function can adaptively learn the binary codes and be trained via back-propagation.
Extensive experiments on three benchmark datasets demonstrate that DBRC outperforms several state-of-the-art methods in both image2text and text2image retrieval task.
\end{abstract}



\maketitle

\section{Introduction}
The same contents or topics can be expressed in multiple kinds of modalities in practice.  For example, the usual speech can be expressed by audio signal or lip movements~\cite{hu2016temporal, hu2016multimodal},
the common content can be described by not only the textual data but also the images~\cite{pang2011multimodal}, and the environment perception could utilize both image and 3D depth information~\cite{wang2014multi}. As these modalities jointly describe the same contents, it becomes possible to make up for each other's limitations and provide more valuable information than single modality.
Hence, there have been many attempts over the years to make use of multimodal data for specific areas, such as audiovisual speech recognition~\cite{hu2016temporal}, image-text classification~\cite{srivastava2012multimodal}.
And the shared contents across modalities also provide possibilities for retrieving relevant data by giving the query of another modality, which has drawn much attention recently~\cite{wang2016comprehensive}.
A typical scenario of such task is to retrieve relevant images by a text query. However, faced with the increasing requirements of massive data organization, storage, and retrieval, traditional cross-modal retrieval shows obvious disadvantages in terms of efficiency.

Recently, hashing method shows its efficiency in approximated nearest neighbor search, which employs the short, binary codes for retrieval instead of the original high dimensional, real-value data.
The binary codes learned from the original database can vastly reduce the storage space and retrieval time. Hence, hashing technique has been widely used in various machine learning and computer vision problems~\cite{weinberger2009feature,kumar2011learning}, especially in unimodal retrieval~\cite{gong2013iterative,li2017large}.
For cross-modal retrieval, hashing has also attracted considerable research attention due to its efficiency.  Cross-modal hashing aims to discover the correlations across modalities to enable the cross-modal similarity search~\cite{wang2016comprehensive}. Hence, different from the unimodal hashing, it should preserve not only the intra-modality consistency, but also the inter-modality consistency.
In this paper, we focus on unsupervised cross-modal hashing technique.

Unsupervised cross-modal hashing problem has been just proposed in recent years. Most of existing hashing methods employ a two-stage framework for learning hashing codes, which first generates the real-value codes in a learned shared semantic space across different modalities, then binarizes the real-value codes via thresholding~\cite{zhou2014latent,ding2016large}.
But such methods are usually based on the shallow model, where linear projection is a common selection for semantic space learning. Hence, the nonlinear correlation across modalities could not be effectively learned. Recently, as the effectiveness of deep networks in producing useful representation has been confirmed~\cite{salakhutdinov2009semantic}, some works choose to learn the common semantic space via a shared layer across the multi-layer nonlinear projection of different modalities~\cite{wang2016comprehensive}.
However, these works based on multimodal networks just provide an empirical analysis in preserving the intra- and inter-modality consistency, which could be unreliable for learning efficient codes.
More importantly, the above works do not directly learn the hashing codes, but are just a simple combination of conventional cross-modal network and binarization. Such frameworks actually relax the binary constraint, and the extra binarization may destroy the learned semantic space and result in a sub-optimal solution. Although Courbariaux~\emph{et al.}~\cite{courbariaux2016binarized} aim to make the weights and activations binary, such model still suffers from difficult optimization.

In this paper, to figure out the effectiveness of multimodal deep network in cross-modal hashing, we provide a theoretical analysis about the usually employed ~\emph{Multimodal Restricted Boltzmann Machine} (MRBM) with \emph{Maximum Likelihood Learning} (MLL). We show that such deep networks with a shared layer across modalities can simultaneously preserve the intra- and inter-modality consistency.
Then, based on the above conclusion, we propose to directly learn the binary hashing codes via a multimodal deep reconstruction network, which is called as \emph{Deep Binary Reconstruction} (DBRC).
In the proposed DBRC, we introduce a novel hashing activation function, named as~\emph{Adaptive Tanh} (ATanh). The hashing layer with ATanh function can adaptively map the activations of previous layers into approximated binary codes. Then, based on the projected hamming semantic space, the original multimodal data is reconstructed in an unsupervised fashion.
The proposed hashing layer makes it possible to simultaneously learn the hashing codes and optimize the deep networks via back-propagation, which could learn more efficient binary codes than the two-stage methods\footnote{The code is available at dtaoo.github.io}.
We conduct extensive experiments on three benchmark datasets, and DBRC shows better codes over state-of-the-art methods on various metrics. 

In the following sections, we first revisit the related cross-modal hashing methods in Section 2. Then we give a concrete analysis about the MRBM with MLL objective in preserving modal consistency in Section 3. In Section 4, we introduce the hashing activation function ATanh, and corresponding optimization method. We then propose DBRC cross-modal hashing framework. Experiments are conducted for evaluation in Section 5. Section 6 concludes this paper.




\section{Related work}
Existing unsupervised cross-modal hashing methods share similar framework for fast retrieval: the data of different modalities are projected into a common low-dimensional space, then binarization operation is performed over the projected real-value vector to obtain binary codes. And these methods can be grouped into two categories with respect to the projection manner: linear modeling and nonlinear modeling methods~\cite{wang2016comprehensive}.

\noindent \textbf{Linear modeling.}
Linear modeling methods aim to utilize linear projection function to learn the common subspace.
\emph{Cross View Hashing} (CVH) ~\cite{kumar2011learning} and \emph{Inter-Media Hashing} (IMH)~\cite{song2013inter} extend the unimodal spectral hashing to multimodal scenario and aim to retain the inter- and intra-modal consistency in the common subspace. Note that \emph{Canonical Correlation Analysis} (CCA) is actually a special case of CVH, which targets to find effective linear projections of different modalities that are maximally correlated. Rastegari \emph{et al.} \cite{rastegari2013predictable} propose \emph{Predictable Dual-view Hashing} (PDH) to refine the CCA projection via ignoring the orthogonal bases and learning the hashing codes in a self-taught manner. Apart from the CCA-like methods, Zhou \emph{et al.}~\cite{zhou2014latent} propose another novel \emph{Latent Semantic Sparse Hashing} (LSSH) for cross-modal hashing.
It aims to maximally correlate the learned latent semantic features of different modalities, which are obtained from sparse coding and matrix factorization. Similarly with LSSH, \emph{Collective Matrix Factorization Hashing} (CMFH)~\cite{ding2016large} assumes that different modalities can be factorized into modality-specific matrices and latent semantic matrix, meanwhile the original modality can be linearly reconstructed from the semantic matrix. And the hashing codes are obtained via a thresholding operation over the common semantic matrix.
These methods are all based on linear modeling that limits their effectiveness in the common subspace modeling.

\noindent \textbf{Nonlinear modeling.}
To overcome the limits of linear modeling, nonlinear modeling based on deep networks has attracted much attention recently.
To our best knowledge, Multimodal Deep Autoencoder (MDAE)~\cite{ngiam2011multimodal} is the first one employing deep networks in multimodal learning.
Concretely, MDAE focuses on audiovisual speech recognition task. It learns the joint representation across modalities via a shared layer of different modality-specific networks, and the whole network is trained by minimizing the reconstruction error of both modalities. Similar frameworks have also been served to multimodal retrieval~\cite{srivastava2012multimodal}.
Recently, there have been some attempts to employ such frameworks for cross-modal hashing problem.
Wang~\emph{et al.}~\cite{wang2015deep} directly employ MDAE network for cross-modal hashing, but impose the orthogonal regularizer on the weights of MDAE to make the codes more efficient. And the hashing codes in \cite{wang2015deep} are obtained by performing an indicator function over the joint representation.
Differently, Feng~\emph{et al.}~\cite{feng2014cross} and Wang~\emph{et al.} \cite{wang2014effective} propose to employ stacked specific-networks to encode each modality, then learn the latent representation by maximizing the semantic similarity of different modalities.
Although these methods make use of the advantages of deep networks in nonlinear modeling, they fail to take consideration of the binary constraint of hashing codes when training the multimodal networks. In other words, they just perform a thresholding operation over the learned joint representation across modalities, which could destroy the original representation and make the codes inefficient.
Unlike the two-stage strategy in these methods, our model can directly generate the hashing codes and perform nonlinear modeling over the modalities.
Moreover, the multimodal networks in these works are designed to preserve the inter- and intra-modal consistency under heuristic consideration, which is not convincing to some extent. Hence, we provide a concrete analysis about such models.

\begin{figure}[b]
\centering
\includegraphics[width=8cm]{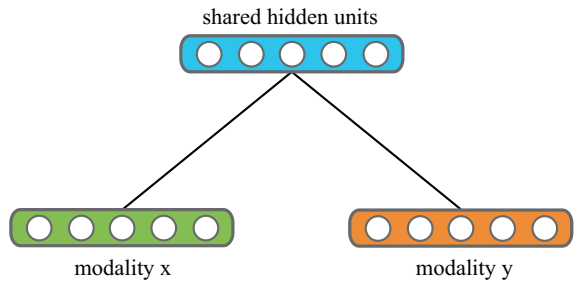}\\
\caption{An illustration of MRBM model.}\label{MRBM}
\end{figure}

\section{Multimodal Maximum Likelihood Learning}
Different from unimodal hashing, cross-modal hashing need to preserve both the inter- and intra-modal correlation.
To capture such correlations, multimodal deep networks propose to fuse multiple modality-specific networks with the help of one shared layer~\cite{ngiam2011multimodal, sohn2014improved, wang2015deep}.
And the fusion scheme is almost always based on~\emph{Multimodal Restricted Boltzmann Machine} (MRBM), which is the ~\textbf{core} of cross-modal retrieval networks, as shown in Fig.~\ref{MRBM}.

MRBM is a special case of RBM that is an energy-based network. To be specific, RBM is an undirected graphical model which defines a probability distribution of visible units using hidden units. Under the case of multimodal input, MRBM defines a joint distribution over modality $\textbf{x}$, modality $\textbf{y}$, and shared hidden units $\textbf{h}$~\cite{srivastava2012multimodal}, which is written as,
\begin{equation} \label{rbm1}
P\left( {\textbf{x},\textbf{y},\textbf{h}} \right) = \frac{1}{Z}\exp \left( { - E\left( {\textbf{x},\textbf{y},\textbf{h}} \right)} \right),
\end{equation}
where $Z$ is the partition function and $E$ is an energy term given by
\begin{equation}\label{rbm2}
\begin{split}
E\left( {\textbf{x},\textbf{y},\textbf{h}} \right) =  - {\textbf{x}^T}\textbf{W}^x\textbf{h} - {\textbf{y}^T}\textbf{W}^y\textbf{h} \rm{~~~~}\\
{\rm{~~~~~~~~}}- {\textbf{x}^T}{\textbf{b}^x} - {\textbf{y}^T}{\textbf{b}^y} - {\textbf{h}^T}{\textbf{b}^h},
\end{split}
\end{equation}
where $\textbf{x}$ and $\textbf{y}$ are the visible units of modality $\textbf{x}$ and $\textbf{y}$, and $\textbf{h}$ is the shared hidden units.
$\textbf{W}^x$ is a matrix of pairwise weights between elements of $\textbf{x}$ and $\textbf{h}$, and similar for $\textbf{W}^y$. $\textbf{b}^x$, $\textbf{b}^y$, and $\textbf{b}^h$ are bias vectors for $\textbf{x}$, $\textbf{y}$, and $\textbf{h}$, respectively.
To obtain the joint likelihood $P(\textbf{x},\textbf{y})$, $\textbf{h}$ is marginalized out from the distribution,
\begin{equation}\label{rbm3}
P(\textbf{x},\textbf{y}) = \sum\limits_\textbf{h} {{{\exp ( - E(\textbf{x},\textbf{y},\textbf{h}))} \mathord{\left/
 {\vphantom {{\exp ( - E(\textbf{x},\textbf{y},\textbf{h}))} Z}} \right.
 \kern-\nulldelimiterspace} Z}}.
\end{equation}

For the MRBM model, similar to the standard RBM, \emph{Contrastive Divergence} (CD) \cite{hinton2006reducing} or \emph{Persistent CD} (PCD) \cite{tieleman2008training} can be used to approximate the gradient to maximize the joint likelihood, i.e., $P(\textbf{x},\textbf{y})$. This is the typical maximum likelihood learning for MRBM.

If let $P_\theta (\textbf{x},\textbf{y})$ denote the MRBM joint distribution parameterized by $\theta = \left\{ \textbf{W}^*, \textbf{b}^* \right\}$ and $P_D (\textbf{x},\textbf{y})$ denote the data generating distribution,  the MLL of MRBM can be re-written as follows,
\begin{eqnarray*}\label{MLL}
  MLL &=&  - {E_{{P_D}\left( {\textbf{x},\textbf{y}} \right)}}\left[ {\log {P_\theta }\left( {\textbf{x},\textbf{y}} \right)} \right]  \\
            &=&  - {E_{{P_D}\left( \textbf{x} \right)}}\left[ {{E_{{P_D}\left( {\textbf{y}|\textbf{x}} \right)}}\log {P_\theta }\left( \textbf{x} \right)} \right]\\
            & &         - {E_{{P_D}\left( \textbf{x} \right)}}\left[ {{E_{{P_D}\left( {\textbf{y}|\textbf{x}} \right)}}\log {P_\theta }\left( {\textbf{y}|\textbf{x}} \right)} \right]\\
            &=& {E_{{P_D}\left( \textbf{x} \right)}}\left[ {{E_{{P_D}\left( {\textbf{y}|\textbf{x}} \right)}}\log \frac{{{P_D}\left( \textbf{x} \right)}}{{{P_\theta }\left( \textbf{x} \right)}}} \right] \\
            & &          + {E_{{P_D}\left( \textbf{x} \right)}}\left[ {{E_{{P_D}\left( {\textbf{y}|\textbf{x}} \right)}}\log \frac{{{P_D}\left( {\textbf{y}|\textbf{x}} \right)}}{{{P_\theta }\left( {\textbf{y}|\textbf{x}} \right)}}} \right] + C\\
            &=& {E_{{P_D}\left( \textbf{x} \right)}}\left[ {{E_{{P_D}\left( {\textbf{y}|\textbf{x}} \right)}}\log \frac{{{P_D}\left( {\textbf{y}|\textbf{x}} \right)}}{{{P_\theta }\left( {\textbf{y}|\textbf{x}} \right)}}} \right] \\
            & &   + {E_{{P_D}\left( \textbf{x} \right)}}\left[ {\log \frac{{{P_D}\left( \textbf{x} \right)}}{{{P_\theta }\left( \textbf{x} \right)}}} \right] + C\\
            &=&   \underbrace {{E_{{P_D}\left( \textbf{x} \right)}}\left[ {KL({P_D}\left( {\textbf{y}|\textbf{x}} \right)||{P_\theta }\left( {\textbf{y}|\textbf{x}} \right))} \right]}_{{\text{cross modalities}}} \\
            & &    +  \underbrace {KL({P_D}\left( \textbf{x} \right)||{P_\theta }\left( \textbf{x} \right))}_{{\text{single modality}}}    + C
\end{eqnarray*}
where $C$ is a constant that is irrelevant to $\theta$. Note that, the above formula can also be re-written with respect to $\textbf{y}$.
It is easy to find that the MLL objective of MRBM consists of two terms: one is related to the distribution of single modality $\textbf{x}$ and the other one is about the conditional probability of cross-modalities.
In other words, maximizing the joint distribution $P_\theta(\textbf{x},\textbf{y})$ is equal to simultaneously learning the unimodal and cross-modal data distribution, which actually preserves both the intra- and inter-modal consistency. Hence, the deep networks based on MRBM have the ability to meet the requirements of cross-modal hashing, and the experimental results of previous works also confirm this.

\section{Deep Binary Reconstruction}
In this section, based on the above concrete analysis, we first propose ATanh activation function for generating binary codes, then show the optimization w.r.t. the function. Finally, a multimodal deep binary reconstruction network is proposed for cross-modal hashing.

\begin{figure}[b]
\centering
\includegraphics[width=8.5cm]{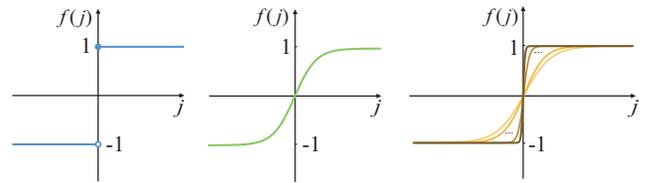}\\
\caption{The comparison among sign, tanh, and ATanh function. The ATanh is initialized by a small value of $\alpha$ (in pale yellow), then adaptively approaching the sign function (in dark yellow).}\label{atanh}
\end{figure}

\begin{figure*}[t]
\centering
\includegraphics[width=16cm]{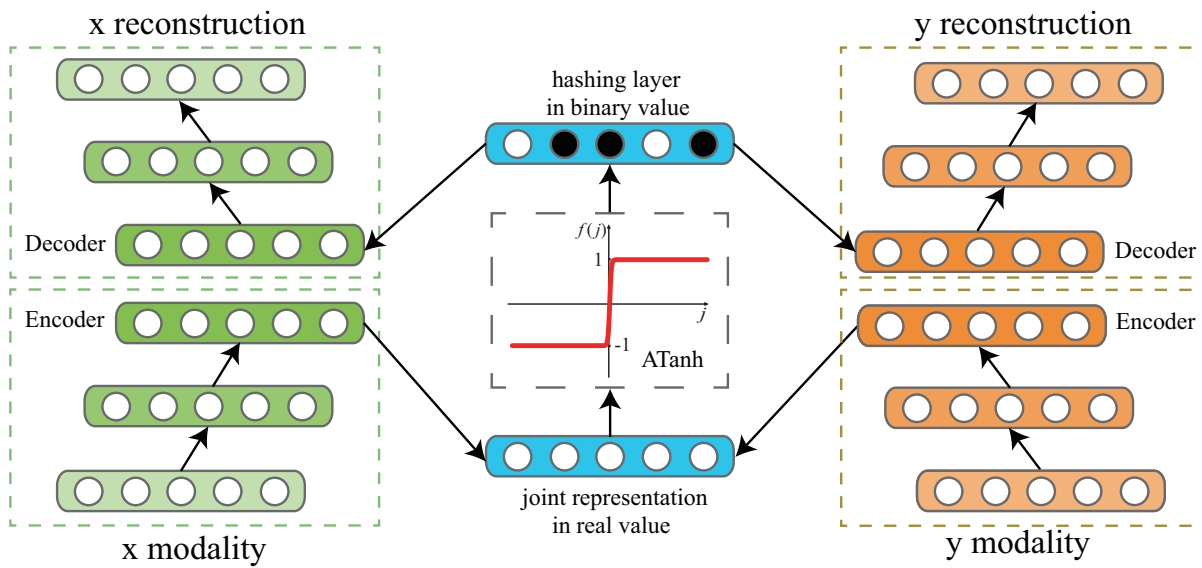}\\
\caption{The proposed deep binary reconstruction network. The joint representation across the two modality-specific networks is adaptively binarized into the hashing layer by performing the ATanh activation. And the whole network is trained to minimize the reconstruction error based on the shared binary representation.}\label{framework}
\end{figure*}

\subsection{Adaptive Tanh}
The hashing technique requires the generated codes to be binary, i.e., $\left\{ { - 1,1} \right\}$. As the binary constraint makes it hard to optimize the networks, conventional two-stage methods perform the sign function over the activations of tanh or sigmoid\footnote{The sign function has an offset of $-0.5$ for the activation values of sigmoid function.} after training the deep networks, as shown in Fig.~\ref{atanh}. In this paper, we propose an adaptive tanh function that is similar to the tanh function but controlled by a learnable scaling parameter $\alpha > 0$, which is defined as,
\begin{equation}\label{tanh1}
f({\textbf{s}}) = \tanh ({\alpha }{\textbf{s}}),
\end{equation}
where $\textbf{s}$ is the activation of previous layers. In Eq.~\ref{tanh1}, it is easy to find that when $\alpha$ is small, especially $\alpha=1$, ATanh becomes tanh, where the activations change gently between $-1$ and $1$.
Conversely, when $\alpha$ is large enough, the proposed ATanh approaches the sign function, which means the activations of ATanh fall into the binary value of hashing codes.
In particular, ATanh is differentiable everywhere, which is totally different from the sign function. Hence, the deep network using ATanh as the activation function can be optimized via back-propagation.

Although the parameter $\alpha$ makes ATanh learnable compared with other functions, it is hard to guarantee that the activations of ATanh fall into the binary codes after training the deep networks.
Actually, $\alpha$ should gradually increase so that the final ATanh has the ability to generate the binary hashing codes. Hence, the activation function becomes
\begin{equation}\label{tanh1_2}
f({\textbf{s}}) = \tanh ({\alpha }{\textbf{s}}) + \lambda  \left\| {\alpha^{ - 1}} \right\|_2^2,
\end{equation}
where $\lambda$ is a regularization constant. The regularization term, $ \left\| {\alpha^{ - 1}} \right\|_2^2$, is a penalty to the original ATanh function, which provides a convenient way to control the magnitude of $\alpha$.
Hence, the new function Eq. \ref{tanh1_2} becomes more reliable for binary code learning.

Note that, the proposed ATanh function is an element-wise function,  so that the introduced $\alpha$ is different for different bits. In other words, we could simultaneously learn 32 ATanh functions for $32~bits$ hashing codes, which makes the codes more adaptable compared with the consistent sign function.

\subsection{Optimization}
As the proposed ATanh is derivable, it can be jointly trained with other layers via back-propagation. Eq. \ref{tanh1_2} can be re-written into element-wise,
\begin{equation}\label{tanh1_3}
f({s_i}) = \tanh ({\alpha _i}{s_i}) + \lambda {\left| {{\alpha _i}} \right|^{ - 2}},  {\rm{~~~~~~}} i = 1,2,...,bits
\end{equation}
where $bits$ is the code length. For each ATanh function $f({s_i})$, the update of ${\alpha}_i$ can be simply derived by chain rule.
Let $\varepsilon$ denote the objective function (e.g., reconstruction error), the partial derivative w.r.t. ${\alpha}_i$ becomes
\begin{equation}\label{tanh2}
\frac{{\partial \varepsilon }}{{\partial {\alpha _i}}} = \frac{{\partial \varepsilon }}{{\partial f({s_i})}}\frac{{\partial f({s_i})}}{{\partial {\alpha _i}}}.
\end{equation}
Here, the first term is the gradient to current hashing layer, which is propagated from previous layers. And the second term is the derivative of ATanh function,
\begin{equation}\label{tanh3}
\frac{{\partial f({s_i})}}{{\partial {\alpha _i}}} = \left( {1 - {\tanh}^2 ({\alpha _i}{s_i})} \right){s_i} -  2\lambda \alpha _i^{ - 3}.
\end{equation}
The update of ${\alpha}_i$ can be performed by employing stochastic gradient descent with RMSprop~\cite{tieleman2012lecture} based on the derivative (Eq. \ref{tanh2}). RMSprop adaptively rescales the step size for updating trainable weights according to the corresponding gradient history.
Note that, when ${\alpha}_i$ becomes larger, it is more easily influenced by the vanishing gradient.
Hence, in the experiments, we follow the empirical parameter setting and ${\alpha}_i = 1$ is considered as the initialization.
And compared with other trainable parameters of the network, the time complexity of ATanh in both forward and backward computation is negligible.

\subsection{Binary Reconstruction Network}
Based on the concrete analysis of the effectiveness of MRBM in preserving the inter- and intra-modal consistency, we propose a novel multimodal \emph{Deep Binary Reconstruction}(DBRC)  network that can directly generate cross-modal hashing codes, as shown in Fig.~\ref{framework}.
Specifically, the high-dimensional data of each modality is first encoded into the low-dimensional representation via modality-specific network, which could capture the data manifold based on the nonlinear modeling of multi-layers~\cite{salakhutdinov2009semantic,do2015discrete}. Then the joint representation across modalities is learned via MRBM model (i.e., Fig.~\ref{MRBM}).
To generate the binary codes within the network, the real-value representation of MRBM is binarized into the hashing codes by taking advantage of the proposed ATanh function, which become the shared hashing layer.
Finally, we can directly reconstruct the original data of each modality based on the binary representation.
As the ATanh function is derivable, the whole network can be trained via back-propagation, and directly generate the embedded binary codes\footnote{Although the proposed ATanh function is very close to the sign function after training, there are still very few activations falling into the interval $\left( {{\rm{ - }}1,1} \right)$. Hence, we simply perform binarization over these activations.}.

As the proposed DBRC model learns the binary representation from both modalities, these modalities share the identical hashing codes.
But it is not suitable for the testing data, as only one modality is available in the retrieval scenario.
Hence, we propose to employ different models for these different scenarios.

\noindent \textbf{Hashing codes for training data.}
As all the modalities are available in the training phase, the hashing codes can be learned by simply training the proposed DBRC model.

\noindent \textbf{Hashing codes for testing data.}
Inspired by Ngiam~\emph{et al.}~\cite{ngiam2011multimodal} and Hu~\emph{et al.}~\cite{hu2016multimodal}, we propose to learn the modality-specific hashing codes by retaining the original value of one modality and setting the other modality to zero when faced with single modality input. In other words, we require the model to reconstruct both modalities with only one modality, just like the video-only deep autoencoder in~\cite{ngiam2011multimodal}. In this case, the model can still preserve the inter- and intra-modal consistency, as the MRBM model is retained in the reconstruction phase. And the original joint representation of MRBM is replaced with the hashing layer.
In practice, we find that such reconstruction model performs better when initialized from complete DBRC model then trained with the unimodal input data.

\section{Experiments}
In this section, we show the results of DBRC compared with other models on three datasets, including image2text and text2image retrieval task. Different code lengths are considered for evaluating the performance. In addition, we also provide an analysis about the sensitivity of hyper-parameter $\lambda$.

\subsection{Setup}
\noindent \textbf{Dataset.} Three benchmark datasets are chosen for evaluation, including Wiki\footnote{http://www.svcl.ucsd.edu/projects/crossmodal/}~\cite{rasiwasia2010new}, FLICKR-25K\footnote{http://press.liacs.nl/mirflickr/}~\cite{huiskes2008mir}, and NUS-WIDE\footnote{http://lms.comp.nus.edu.sg/research/NUS-WIDE.htm}~\cite{chua2009nus}.

\noindent \textbf{Wiki} is an image-text dataset, which is collected from Wikipedia's ``featured article". There are 2,866 pairs in the dataset. For each pair, the image modality is represented as  128-dimensional SIFT descriptor histograms, and text is expressed as 10-dimensional semantic vector via latent Dirichlet allocation model. These pairs are annotated with one of 10 topic labels.
In this paper, we choose 25\% of the dataset as the query set and the rest for retrieval set.

\noindent \textbf{FLICKR-25K} is an image collection from Flickr, where 25,000 images are associated with multiple textual tags (text).  The average number of tags for each image is about 5.1~\cite{srivastava2012multimodal}. And these image-tag pairs are annotated by 24 provided labels. Following the setting in \cite{lin2015semantics}, we select the textual tags which appear more than 20 times and keep the valid pairs. The left images are represented with 150-dimensional edge histogram and the texts are expressed as 500-dimensional tagging vector.
Here we  take 5\% of the dataset as the query set and the rest for training set.

\noindent \textbf{NUS-WIDE} dataset consists of 269,648 multi-label images. Each image is also associated with multiple tags (6 in average). The image-tag pairs are annotated with 81 concept labels.
Among these concepts, the common 10 ones are considered in our experiments. The images are represented into 500-dimensional bag-of-words based on SIFT descriptor. The textual tags are expressed with 1000-dimensional tag occurrence vector. 4000 image-tag pairs are uniformly sampled as the query set, and the rest ones are served as the training set.

\noindent \textbf{Evaluation.}
In this paper, we focus on two cross-modal retrieval task, i.e., image2text (I2T) and text2image (T2I).
Hamming ranking and hash lookup are both employed for evaluation. Specifically, \emph{Mean Average Precision} (MAP) is computed based on the Hamming distance to a query for Hamming ranking, and the hash lookup performance is according to a Hamming ball of radius 2 to a query. And the ground-truth of relevant items for a query are defined as whether they share at least one common label.

\begin{table*}[t]
\centering
\newcolumntype{C}[1]{>{\centering}p{#1}}
\renewcommand\arraystretch{1.2}
\small
\begin{tabular}{c|C{2.2cm}|C{0.8cm}C{0.8cm}C{0.8cm}C{0.8cm}|C{0.8cm}C{0.8cm}C{0.8cm}C{0.8cm}|C{0.8cm}C{0.8cm}C{0.8cm}C{0.8cm}}
\hline
{\multirow{2}*{Task}} &   Dataset &  \multicolumn{4}{c|}{Wiki} &  \multicolumn{4}{c|}{FLICKR-25K}  &   \multicolumn{4}{c}{NUS-WIDE}
\tabularnewline\cline{2-14}

& Code Length & $16~bits$ & $32~bits$ & $64~bits$ &  $128~bits$ & $16~bits$ & $32~bits$ & $64~bits$ &  $128~bits$ & $16~bits$ & $32~bits$ & $64~bits$ &  $128~bits$
\tabularnewline
\hline
\hline
\multirow{7}*{I2T}& IMH~\cite{song2013inter}                        &   0.1593  & 0.1477  & 0.1420  & 0.1291         & 0.5621  & 0.5643  & 0.5649  & 0.5642         & \textbf{0.4187}  & 0.3975  & 0.3778  & 0.3668  \tabularnewline

& CVH~\cite{kumar2011learning}                       &   0.1993  &  0.1889  & 0.1803  & 0.1782         & 0.5815  & 0.5756  & 0.5710  & 0.5677         & 0.3888  & 0.3744  & 0.3621  & 0.3537  \tabularnewline

& CMFH~\cite{ding2016large}                      &   0.2126  & 0.2208  & 0.2322  & 0.2337         & 0.5721  & 0.5740  & 0.5739  & 0.5736         & 0.3443  & 0.3438  & 0.3454  & 0.3461  \tabularnewline

& LSSH~\cite{zhou2014latent}                      &   0.2122  & 0.2260  & 0.2155  & 0.2297         & 0.5779  & 0.5795  & 0.5848  & 0.5878         & 0.3891  & 0.3910  & 0.3977  & 0.3949  \tabularnewline

& Corr-Full-AE~\cite{feng2014cross}        &   0.1802  & 0.1937  & 0.1911  & 0.2014         & 0.5557  & 0.5551  & 0.5583  & 0.5553         & 0.3468  & 0.3468  & 0.3470  & 0.3410   \tabularnewline

& DMHOR~\cite{wang2015deep}                &   0.1919  & 0.1841  & 0.1847  & 0.1877         & 0.5848  & 0.5810  & 0.5842  & 0.5851         & 0.3657  & 0.3620  & 0.3678  & 0.3590  \tabularnewline

& DBRC                    &   \textbf{0.2534}  & \textbf{0.2648}  & \textbf{0.2686}  & \textbf{0.2878}         & \textbf{0.5873 } & \textbf{0.5898 } & \textbf{0.5902 } & \textbf{0.5907}         & 0.3939  & \textbf{0.4087}  & \textbf{0.4166}  & \textbf{0.4165}  \tabularnewline

\hline
\hline

\multirow{7}*{T2I}& IMH~\cite{song2013inter}                        & 0.1417  & 0.1297  & 0.1243  & 0.1105         & 0.5624  & 0.5643  & 0.5651  & 0.5648         & 0.4053  &  0.3892  &0.3758   & 0.3627  \tabularnewline

& CVH~\cite{kumar2011learning}                       &  0.1652  & 0.1582  & 0.1512  & 0.1469         & 0.5817  & 0.5761  & 0.5715  & 0.5681         & 0.3822  &  0.3697  &0.3592  & 0.3519    \tabularnewline

& CMFH~\cite{ding2016large}                    & 0.4830  & 0.5147  & 0.5338  & 0.5370         & 0.5673  & 0.5693  & 0.5681  & 0.5682         & 0.3506  &  0.3509  &0.3524  & 0.3547    \tabularnewline

& LSSH~\cite{zhou2014latent}                      & 0.4992  & 0.5245  & 0.5326  & 0.5395         & 0.5874  & 0.5926  & 0.5957  & 0.5964         & 0.4115  & 0.4162   &0.4229  & 0.4198    \tabularnewline

& Corr-Full-AE~\cite{feng2014cross}        &  0.1410  & 0.1262  & 0.1366  & 0.1483         & 0.5576  & 0.5545  & 0.5576  & 0.5567         & 0.3385  &  0.3438  &0.3390  & 0.3382    \tabularnewline

& DMHOR~\cite{wang2015deep}                & 0.4272  & 0.4874  & 0.4916  & 0.4818         & 0.5664  & 0.5622  & 0.5540  & 0.5653         & 0.3724  &  0.3613  &0.3498  & 0.3401    \tabularnewline

& DBRC                    & \textbf{0.5439}  & \textbf{0.5377}  & \textbf{0.5476}  & \textbf{0.5488 }        & \textbf{0.5883}  & \textbf{0.5963}  & \textbf{0.5962}  & \textbf{0.5975}         & \textbf{0.4249}  &  \textbf{0.4294}  &\textbf{0.4381}  & \textbf{0.4427}    \tabularnewline
\hline
\hline
\end{tabular}
\caption{\label{Table}Hamming ranking performance (in MAP) on Wiki, FLICKR-25K, and NUS-WIDE dataset with varying code lengths.}
\end{table*}

\noindent \textbf{Baselines.}
The proposed method is compared with several unsupervised cross-modal hashing methods, including IMH~\cite{song2013inter}, CVH~\cite{kumar2011learning}, CMFH (UCMFH)~\cite{ding2016large}, LSSH~\cite{zhou2014latent}, Corr-Full-AE~\cite{feng2014cross}, and DMHOR~\cite{wang2015deep}.
Note that, the first four are based on linear modeling, while the last two are based on nonlinear modeling (deep networks).
Source codes of IMH, CVH, CMFH, and LSSH are provided by the corresponding authors. While the rest two are not available, so we implement them carefully.
For DMHOR, we follow the network architecture and hyper-parameter settings introduced in the original paper. While for Corr-Full-AE, the detailed network settings are not provided, so we choose the similar network as DBRC for fairness. Note that, the initialization and optimization method are also not provided,  so we try different strategies for training  Corr-Full-AE. However, it still tends to map all the original data into similar codes.
Hence, we only compare it with DBRC in Hamming ranking.

\subsection{Model architecture}
The proposed DBRC model consists of two pathways for the two modalities (i.e., image and text). The image pathway consists of an encoder and a decoder, where the encoder is a 3-layers networks  ($n$-128-512, $n$ is the unit number of input feature), while the decoder takes 512-128-$n$ settings. And we take the same networks for the text modality.
Note that, due to the efficient gradient propagation of ReLu, it is chosen as the activation function of DBRC except the hashing layer and the joint layer of MRBM.
The hyper-parameter of $\lambda$ is set to 0.001 for all the datasets, and we also provide a discussion about it in the following section.

\subsection{Experimental Results}
\begin{figure*}[t]
\centering
\includegraphics[width=17.5cm]{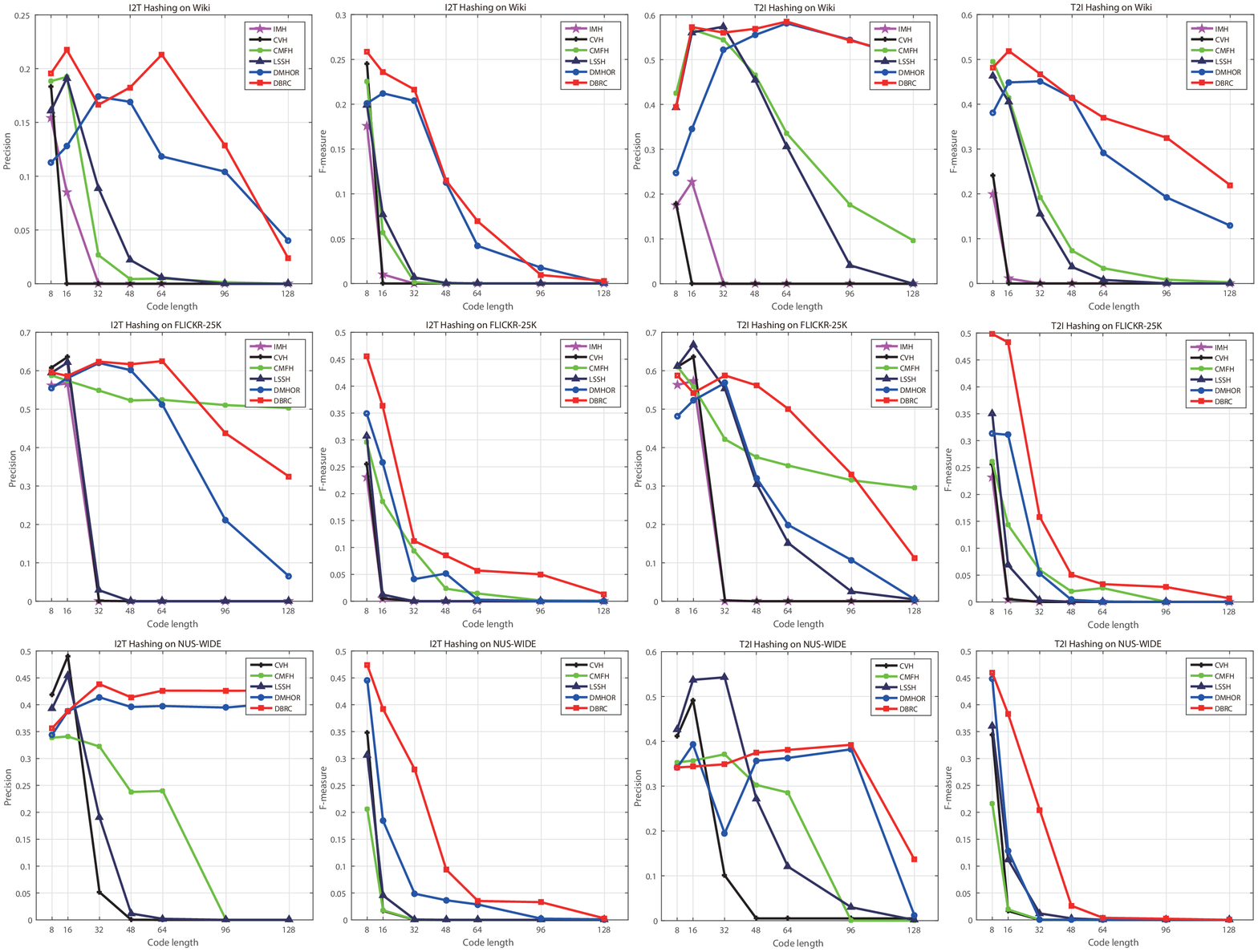}\\
\caption{The hash lookup performance (in Precision and F-measure) of different cross-modal hashing methods on Wiki, FLICKR-25K, and NUS-WIDE dataset with varying code lengths.}\label{lookup}
\end{figure*}

\noindent \textbf{Results on Wiki.}
Table \ref{Table} shows the comparison results on MNIST dataset in MAP (for Hamming ranking).
We can easily find that our proposed DBRC model shows remarkable performance compared with other methods, especially the ones based on deep network.
And we also find that the nonlinear modeling methods (i.e., Corr-Full-AE and DMHOR) do not perform better than conventional linear modeling, especially the state-of-the-art method of CMFH.
This is because these methods based on deep network simply perform the thresholding operation over the hidden units, while the hidden units suffer from the unbalanced activations~\cite{salakhutdinov2009semantic}, which results in inefficient codes. However, DBRC can overcome such weakness by directly learning the binary codes from reconstruction.
The hash lookup results in precision and f-measure are shown in Fig. \ref{lookup}. Different from the comparison results in MAP, the nonlinear modeling methods significantly improve the performance over linear modeling ones. The reason is that hash lookup just focuses on the top retrieved items, i.e., the ones within specific Hamming ball, while hashing ranking relies on the whole retrieved items. In other words, the nonlinear modeling ones can provide more exact results.
Hence, the results in Fig. \ref{lookup} demonstrate that our method can generate more efficient hashing code for cross-modal retrieval, especially when compared with other nonlinear modeling methods.

\noindent \textbf{Results on FLICKR-25K.}
We show the Hamming ranking performance in Table \ref{Table}. It is obvious that DBRC takes the best performance among all the methods.
Corr-Full-AE and DMHOR still suffer from the same problem in Wiki dataset.
And Fig.~\ref{lookup} shows the hash lookup results.
We can find that although most methods decrease sharply with the increasing code length, which results from more sparse hamming space, DBRC still remains substantial superiority over them.
Moreover, CMFH outperforms our method in precision, but DBRC takes better balance between precision and recall and shows superiority over all the other methods.

\noindent \textbf{Results on NUS-WIDE.}
In Table \ref{Table}, DBRC shows the best MAP scores in both image2text and text2image retrieval task.
Here, we report the MAP score of IMH in \cite{wang2016comprehensive} due to the limited memory of our desktop PC. As the hash lookup performance is not provided in the corresponding paper, we do not compare with IMH on NUS-WIDE in Fig.~\ref{lookup}.
In Fig.~\ref{lookup}, the linear modeling methods enjoy better performance in precision at short codes, but decrease rapidly with the increasing code length. While DBRC shows stable performance when faced with longer code length, which shows its effectiveness in learning binary codes in more sparse hamming space.
More importantly, DBRC shows remarkable superiority in F-measure, which also confirms its ability in recalling similar items of another modality.

\noindent \textbf{Parameter sensitivity.}
\begin{figure}[t]
\centering
\includegraphics[width=8cm]{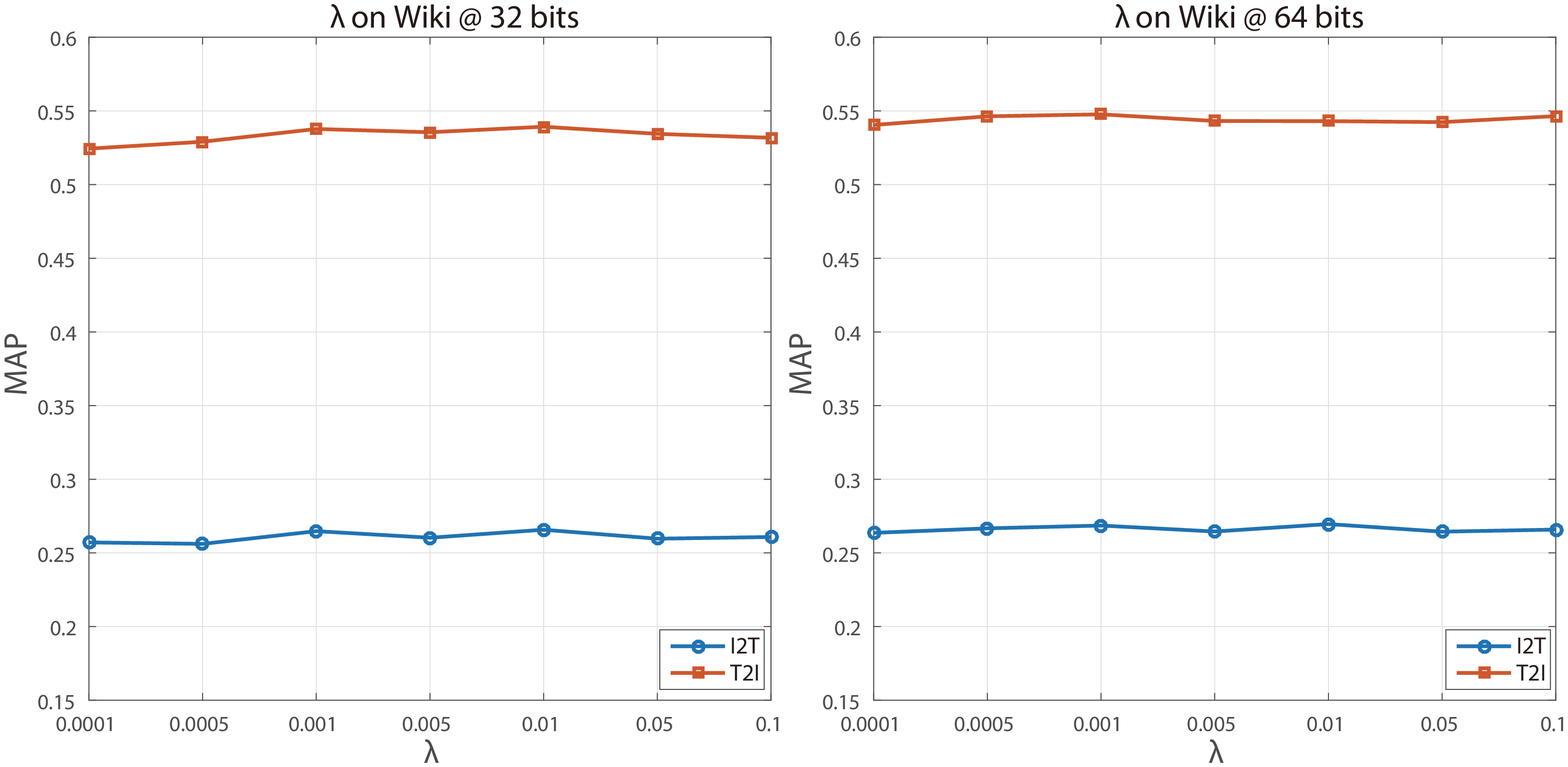}\\
\caption{Parameter sensitivity on Wiki with different code lengths. The results with various settings of $\lambda$ are in MAP scores. }\label{alpha}
\end{figure}
We provide an empirical analysis about the affects of the hyper-parameter of $\lambda$.
Fig. \ref{alpha} shows the results on Wiki dataset with 32 and 64 bits codes (The other two datasets have similar performance).
We can find that DBRC is very insensitive to the choices of $\lambda$ in all the cases, which controls the tradeoff between the adaptive activation and the regularization.
Actually, when $\lambda$ becomes larger, the units in hashing layer tend to be closer to binary value in the initial training stages, which may damage the abstract representation across modalities, just like the sign function.
Hence, DBRC with larger $\lambda$ need more times to converge. In this paper, we choose $\lambda = 0.001$ for all the experiments, which shows remarkable performance over other methods.

\begin{table}[t]
\centering
\newcolumntype{C}[1]{>{\centering}p{#1}}
\renewcommand\arraystretch{1.2}
\small
\begin{tabular}{c|C{1.2cm}|C{0.8cm}C{0.8cm}C{0.8cm}C{0.8cm}C{0.8cm}}
\hline
Task&  Method & $8~bits$ & $16~bits$ & $48~bits$ &  $64~bits$ & $96~bits$
\tabularnewline
\hline
\hline
\multirow{3}*{I2T}& DBRC-C       & 0.2219 & 0.2199 & 0.2234 & 0.2379 & 0.2483       \tabularnewline
&  DBRC-N                                             &  0.2308 & 0.2500 & 0.2616 & 0.2565 & 0.2632 \tabularnewline
& DBRC                    &  0.2327 & 0.2534 & 0.2674 &0.2686 & 0.2736   \tabularnewline
\hline
\hline
\multirow{3}*{I2T}& DBRC-C                        & 0.4342 & 0.5165 & 0.5419 & 0.5351 & 0.5424       \tabularnewline
&  DBRC-N                          &  0.4650 & 0.5382 & 0.5508 & 0.5336 & 0.5469    \tabularnewline
& DBRC                    &   0.4868 & 0.5439 & 0.5538 & 0.5476 & 0.5520  \tabularnewline
\hline
\hline
\end{tabular}
\caption{\label{Table2}Hamming ranking performance (in MAP) with varying code length. Different variants of ATanh are compared on Wiki dataset.}
\end{table}

\noindent \textbf{Activation comparison.}
Here, we make a comparison among different variants of the proposed ATanh activation function.
To address the complex optimization of sign function, DBRC-C~\cite{cao2017hashnet} makes use of  a fixed sequence of $\alpha$ for training the networks one by one.
We also consider the activation function without the regularization term, i.e., Eq.~\ref{tanh1}, which is named as DBRC-N.
As shown in Table.~\ref{Table2}, the learnable activation functions (i.e., DBRC-N and DBRC) show better performance than the fixed one in the two tasks.
This is because they can adaptively learn the binarization function based on the projected low-dimensional subspace, especially for each bit.
While DBRC-C just performs the hard-threshold function over the real-value representation without consideration of the data structure.
On the other hand, DBRC is better than DBRC-N in different code lengths. The reason is that the extra binarization can almostly be ignored in DBRC, when compared with DBRC-N.
Hence, DBRC can learn more effective binary projection in the hashing layer.

\section{Conclusion}
In this paper, we propose to directly learn the cross-modal hashing codes by reconstructing the original data from embedded shared binary representation, which is distinctly different from previous works.
Our model fuses the original two-stage methods and can generate more efficient codes.
And the proposed ATanh activation function gives rise to such superiority, which can adaptively learn the binary codes within networks and be trained via back-propagation.
More importantly, we provide a concrete analysis about the effectiveness of multimodal networks in preserving inter- and intra-consistency for cross-modal retrieval.
Extensive experimental results on three benchmark datasets demonstrate that our proposed method can generate better compact codes in both image2text and text2image retrieval task.

\bibliographystyle{ACM-Reference-Format}
\bibliography{dbrc}

\end{document}